\pdfoutput=1

\documentclass[11pt]{article}

\usepackage[final]{ACL2023}

\usepackage{times}
\usepackage{latexsym}
\usepackage[most]{tcolorbox}
\usepackage{float}

\usepackage[T1]{fontenc}

\usepackage[utf8]{inputenc}

\usepackage{microtype}

\usepackage{inconsolata}

\usepackage{lipsum}
\usepackage{kotex}

\usepackage{babel}
\usepackage{adjustbox}
\usepackage{booktabs}
\usepackage{multirow}
\usepackage{float}

\usepackage{amsmath}
\DeclareMathOperator*{\argmax}{argmax}
\usepackage{amssymb}
\usepackage{bm}

\title{Guard Models Are Overconfident Where Base Models Are Uncertain}

\author{
  \textbf{Jonghyun Hong}\textsuperscript{1} \qquad
  \textbf{MinJae Jung}\textsuperscript{1} \qquad 
  \textbf{Minwoo Kim}\textsuperscript{1}\thanks{\ \ Corresponding author.} \\[0.5em]
  \textsuperscript{1}\,DATUMO INC. \\[0.2em]
  \texttt{\{jonghyun.hong,minjae.jung,mwkim\}@selectstar.ai}
}

\begin{document}
\maketitle
\begin{abstract}
Guard models are used as safety classifiers, with confidence scores driving downstream moderation decisions. We evaluate five guard models for prompt classification and find that although several are nearly calibrated on clean inputs, adversarial attacks degrade their calibration by an order of magnitude, turning false negatives into high-confidence errors indistinguishable from correct detections. Comparing each guard with its corresponding base LM, we find that uncertainty signals often remain available, with the base model typically expressing uncertainty on the same inputs where the guard fails. Layer-wise analyses localize this guard–base divergence to later layers, where guard models exhibit sharper safe/unsafe separation and lower-rank representations, while adversarial harmful inputs lie closer to the clean-safe region. These findings highlight a mismatch between guard confidence and base-model uncertainty under attack.
\textcolor{red}{\textbf{WARNING:} This paper contains model outputs that can be offensive in nature.}
\end{abstract}

\begin{figure*}[t]
    \centering
    \includegraphics[width=\textwidth]{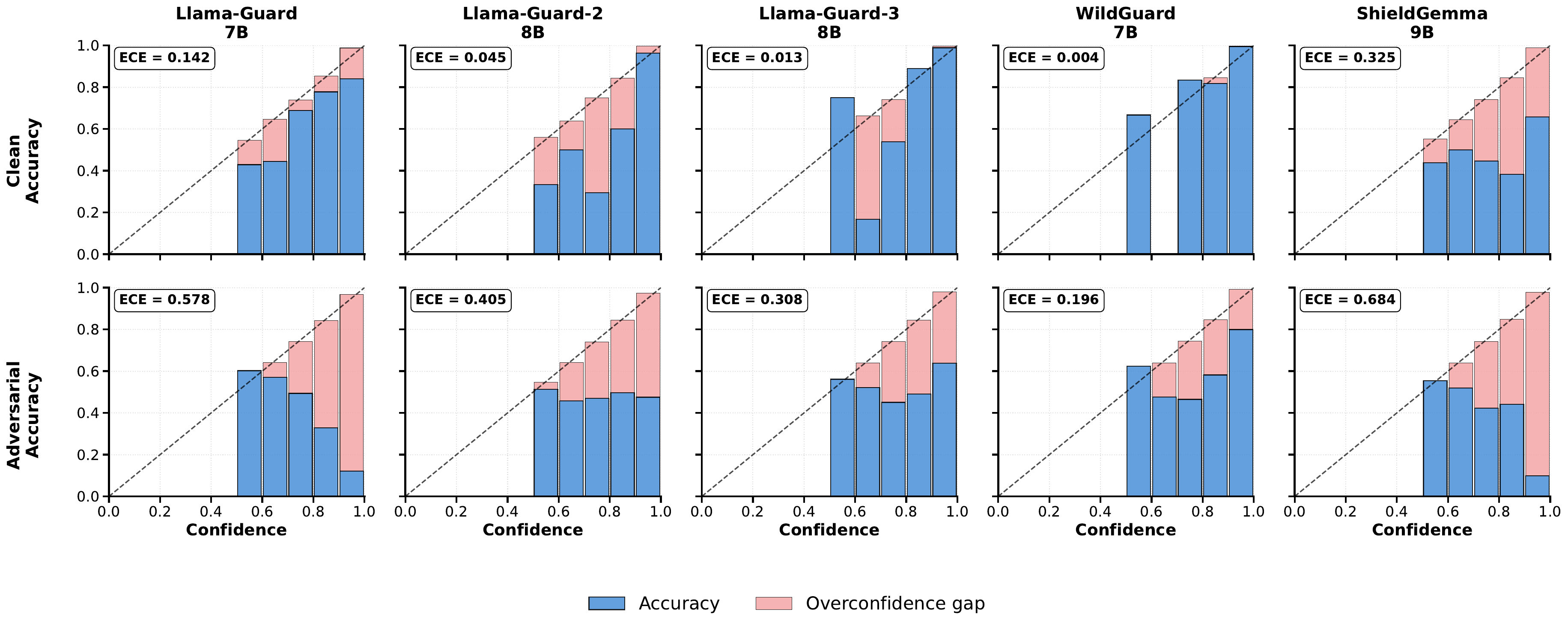}
    \caption{Reliability diagrams for guard model confidence on clean (Top) and adversarial (Bottom) inputs. Blue bars show empirical accuracy per confidence bin, the dashed diagonal denotes perfect calibration, and red regions indicate overconfidence. Several guards are well calibrated on clean inputs, but adversarial inputs degrade calibration by an order of magnitude, with overconfidence concentrated in the highest-confidence bins. A per-attack breakdown is provided in Appendix \ref{sec:Ablation Studies}.}
    \label{fig:main}
\end{figure*}

\section{Introduction}

Large Language Models (LLMs) have become a central interface for user-facing AI systems, building on broad task-solving capabilities and advances in instruction following \citep{brown2020language, ouyang2022training, openai2024gpt4}. 
As their use expands across general-purpose and domain-specific applications, broader deployment also amplifies safety risks, including misuse, manipulation, and jailbreak attacks that induce harmful generations \citep{weidinger2021ethical, ganguli2022red, zou2023universal}. 
Consequently, ensuring the safety and reliability of deployed LLM systems has become a major challenge.

To mitigate these risks, guardrails have increasingly been adopted as external safety layers that operate around the base model at inference time \citep{rebedea2023nemo,yuan2024rigorllm}. These guardrails monitor user inputs and model outputs to enforce policy-specific safety constraints. In many deployment pipelines, this role falls to guard models: auxiliary classifiers that judge prompts or responses against predefined safety taxonomies and decide whether to allow, flag, or block an interaction \citep{inan2023llama, ghosh2024aegis, han2024wildguard, zeng2024shieldgemma}. In practice, these decisions often rely on confidence-based thresholding rather than binary verdicts alone.

Despite their growing role, the reliability of the confidence scores driving these moderation decisions remains underexplored. Guard models inherit the broader calibration problem observed in LLMs, where confidence is often misaligned with actual correctness \citep{kadavath2022language, lin2022teaching, tian2023just, xiong2024can, leng2025taming}. Recent work confirms that this issue extends to guard models, showing severe overconfidence under jailbreak inputs, which post-hoc methods such as temperature scaling, contextual calibration, and batch calibration can partially address \citep{liu2025calibration}, while complementary work has shown that guard predictions can be unstable even under meaning-preserving paraphrases, motivating self-supervised consistency training for semantic robustness \citep{pinneri2025guarding}. In contrast, we diagnose adversarial miscalibration through matched guard–base comparisons of uncertainty and internal representations. Our results indicate that guard failures reflect surface-form biases in safety judgments rather than poor calibration alone, leaving guards vulnerable to jailbreak attacks that disguise harmful intent.

\begin{figure*}[t]
    \centering
    \includegraphics[width=\textwidth]{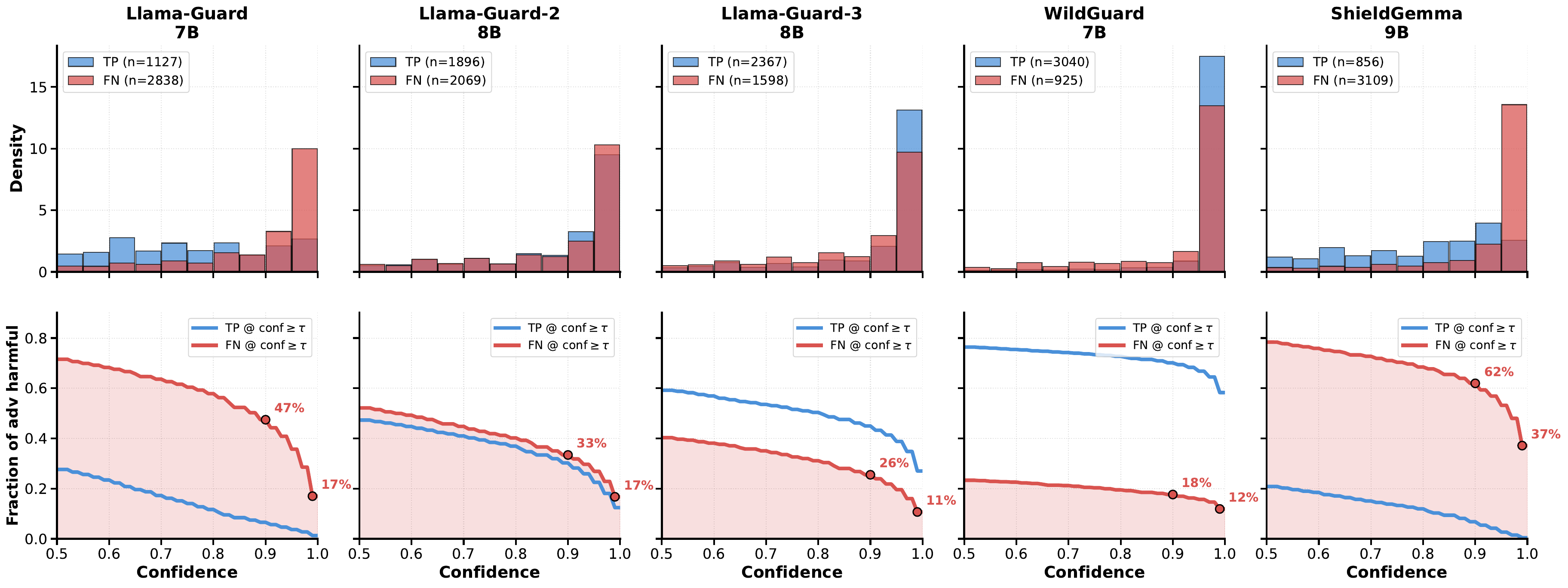}
    \caption{Guard overconfidence persists at every confidence level. 
    (Top) Confidence distributions of true-positive (blue) and 
    false-negative (red) decisions on the pooled adversarial harmful set. 
    For all five guards, FN density concentrates at the high-confidence 
    end, overlapping heavily with TP. (Bottom) Fraction of the 
    adversarial pool retaining a TP or FN decision at confidence 
    $\geq \tau$. Even at $\tau = 0.99$, substantial FN fractions remain, 
    indicating that confidence thresholding cannot selectively filter 
    missed detections.}
    \label{fig:tp_fn}
\end{figure*}

In this work, we evaluate five guard models across six attack methods spanning gradient-based,
template-based, and LLM-based strategies, including
a custom attack method. As Figure~\ref{fig:main} shows, guards that are nearly well calibrated on clean inputs degrade by an order of magnitude under adversarial inputs, with the overconfidence concentrated in high-confidence bins. 
Moreover, Figure~\ref{fig:tp_fn} reveals that false negatives are not low-confidence borderline cases but high-confidence errors. Even at a 0.99 threshold, a substantial fraction of adversarial harmful prompts
pass through as confident false negatives, indistinguishable from correct detections. This raises the question of whether such confident guard failures reflect ambiguity in the adversarial inputs or whether uncertainty remains available in the corresponding base model.

Motivated by findings that pre-trained base LLMs
retain well-calibrated uncertainty even as
post-training erodes it~\citep{nakkiran2025trained,
leng2025taming, luo2025your}, we compare each guard
with its corresponding base LM. Both are evaluated
through a shared binary safe/unsafe interface, with
base LMs using a 5-shot classification prompt. 
We find that adversarial wrapping induces larger verdict shifts in guards than in their corresponding base LMs (Figure~\ref{fig:as}), indicating greater sensitivity to surface-level context changes despite unchanged harmful content. 
Yet this sensitivity is not reflected in uncertainty: on most guard false negatives, the base model assigns higher entropy to the same input (Figure~\ref{fig:ent}). We then localize this divergence with layer-wise and representation analyses. Guards increasingly assign lower unsafe probability than their base models in late layers (Figure~\ref{fig:layerwise}), while their hidden states become concentrated in a lower-rank space with a sharper safe/unsafe separation (Figures~\ref{fig:pca} and~\ref{fig:rank}). This structure separates clean inputs, but often places adversarial harmful inputs near the clean-safe side, producing confident misclassifications.

We make three main contributions:
\begin{itemize}
\item We show that guard model confidence, even when well calibrated on clean inputs, becomes overconfident under adversarial attack, with false negatives reaching confidence levels comparable to correct detections.

\item We compare guards with their corresponding base LMs and find that, on most guard false negatives, base models assign higher entropy to the same input, revealing a systematic uncertainty gap between guards and their corresponding base models.

\item We provide layer-wise evidence that guard–base divergence emerges in later layers, where guards exhibit lower unsafe probabilities, lower-rank representations, and sharper safe/unsafe separation, while adversarial harmful inputs lie closer to the clean-safe region.

\end{itemize}

\section{Related Work}

\paragraph{Guardrails and Guard Models for LLM Safety.}

Content moderation systems such as Perspective API \citep{lees2022new} and
OpenAI's moderation API \citep{markov2023holistic} detect toxic content in
human-generated text but do not address the interactive setting of LLM
deployment, where jailbreak attempts and unsafe generations introduce distinct
risks.
LLM-specific guard models such as Llama-Guard \citep{inan2023llama}, Aegis \citep {ghosh2024aegis}, MD-Judge \citep{li2024salad}, WildGuard \citep{han2024wildguard}, and ShieldGemma \citep{zeng2024shieldgemma} classify prompts or responses against safety taxonomies, evaluated by benchmarks such as HarmBench \citep{mazeika2024harmbench}, AdvBench \citep{zou2023universal}, ToxiGen \citep{hartvigsen2022toxigen}, and RigorLLM \citep{yuan2024rigorllm}. 
System-level frameworks such as NeMo Guardrails \citep{rebedea2023nemo} and SafeRoute \citep{lee2025saferoute} address deployment routing and efficiency.  
This prior work focuses on whether guards correctly detect harmful content, not whether their confidence in those decisions is trustworthy.

\paragraph{Calibration and uncertainty in LLMs.}
LLMs can estimate their own correctness 
\citep{kadavath2022language, lin2022teaching, tian2023just, 
xiong2024can}, yet post-training systematically inflates confidence 
\citep{leng2025taming}. In contrast, pre-trained base LLMs appear to 
retain well-calibrated uncertainty: \citet{nakkiran2025trained} show 
that semantic calibration emerges as a byproduct of next-token 
pre-training and persists under few-shot prompting, while 
\citet{luo2025your} demonstrate that pre-trained representations can 
serve as unsupervised calibrators for post-trained models. Building 
on this insight, BaseCal \citep{tan2026basecal} projects post-trained 
hidden states into the base model's representation space, and 
Dual-Align \citep{luo2026unlocking} corrects layer-wise inference 
drift between pre- and post-trained models. Our work shares the 
premise that base models provide a useful calibration reference but 
uses the comparison diagnostically rather than proposing a correction 
method. These findings also motivate our use of a 5-shot interface 
to elicit base-model uncertainty as a reference.

\section{Background}

\paragraph{Guard models and restricted softmax.}
A guard model is a safety classifier built on top of a language model.
Given a user prompt $x$, it predicts a binary safety label
$\hat{y}(x) \in \{\texttt{safe}, \texttt{unsafe}\}$. We focus on prompt classification, where the guard determines whether the user
input itself should be allowed or blocked before it is passed to a
downstream model. In deployed guard models such as Llama-Guard,
WildGuard, and ShieldGemma, this classification is implemented through
autoregressive generation: the model is prompted with an instruction or
safety policy and generates a label token such as \texttt{safe}/\texttt{unsafe} or an equivalent allow/block label.
We compute the guard's safety distribution by applying a restricted
softmax over the two allowed safety labels. Let $z_{\texttt{safe}}(x)$ and
$z_{\texttt{unsafe}}(x)$ denote the logits assigned to the two labels at
the decision position. Then
\begin{equation}
p_{\texttt{unsafe}}(x)
=
\frac{
\exp z_{\texttt{unsafe}}(x)
}{
\exp z_{\texttt{safe}}(x) + \exp z_{\texttt{unsafe}}(x)
}.
\end{equation}
We set $p_{\texttt{safe}}(x)=1-p_{\texttt{unsafe}}(x)$.
The predicted label is
\begin{equation}
\hat{y}(x)
= \argmax_{y \in \{\texttt{safe}, \texttt{unsafe}\}} p_y(x).
\end{equation}
This restricted distribution measures confidence in the binary safety decision rather than the model's absolute probability mass over the full vocabulary.

\paragraph{Accuracy, confidence, and calibration.}
Accuracy measures whether the predicted label matches the ground-truth
safety label. Clean benign prompts are labeled \texttt{safe}, while clean
harmful prompts are labeled \texttt{unsafe}. For adversarial prompts, the underlying harmful behavior remains \texttt{unsafe} even when it is wrapped by a jailbreak attack.

We define confidence as the probability assigned to the predicted label,
\begin{equation}
c(x)
=
p_{\hat{y}(x)}(x)
=
\max_{y \in \{\texttt{safe},\texttt{unsafe}\}} p_y(x).
\end{equation}
A model is calibrated when its confidence matches its empirical accuracy:
among examples assigned confidence near $p$, approximately a fraction $p$
should be correct. We quantify calibration with Expected Calibration Error
(ECE). Given $B$ confidence bins, let $S_b$ be the set of examples whose
confidence falls in bin $b$. Then
\begin{equation}
\mathrm{ECE}
=
\sum_{b=1}^{B}
\frac{|S_b|}{n}
\left|
\mathrm{acc}(S_b)-\mathrm{conf}(S_b)
\right|,
\end{equation}
where $\mathrm{acc}(S_b)$ is the empirical accuracy in bin $b$ and
$\mathrm{conf}(S_b)$ is the mean confidence. Lower ECE indicates better
calibration.

\section{Guard Confidence Under Adversarial Attack}

We analyze guard model confidence under adversarial inputs in three stages. 
After describing the experimental setup (Section \ref {sec:exp_setup}), we show that guard confidence becomes severely overconfident under attack (Section \ref {sec:guard_attack}). 
We then compare guards with their base LMs to test whether uncertainty signals remain available before safety fine-tuning (Section \ref{sec:base-uncertain}), and finally localize guard--base divergence in the network through layer-wise analyses (Section \ref{sec:layerwise}).

\subsection{Experimental Setup}
\label{sec:exp_setup}

We evaluate guard model confidence by comparing each fine-tuned guard with its corresponding base LM on clean and adversarial prompt-classification inputs.

\paragraph{Models.}

The guards are Llama-Guard 7B/Llama-Guard-2 8B/Llama-Guard-3 8B \citep{inan2023llama}, WildGuard 7B \citep{han2024wildguard}, and ShieldGemma 9B \citep{zeng2024shieldgemma}. Their corresponding base LMs are Llama-2 7B \citep{touvron2023llama}, Llama-3 8B, Llama-3.1 8B \citep{grattafiori2024llama}, Mistral 7B \citep{jiang2023mistral} and Gemma-2 9B \citep{team2024gemma}, respectively.

\paragraph{Data.}

The clean pool contains 713 harmful prompts from HarmBench \citep{mazeika2024harmbench} and StrongREJECT \citep{souly2024strongreject}, and 500 benign prompts from Alpaca \citep{taori2023alpaca}, yielding $1{,}213$ clean prompts. 
To build the adversarial pool, we apply jailbreak attacks to the harmful prompts: GCG \citep{zou2023universal}, AutoDAN \citep{liu2023autodan}, TAP \citep{mehrotra2024tap}, PAIR \citep{chao2023jailbreaking}, AutoDAN\nobreakdash-Turbo \citep{liu2024autodanturbo}, and a custom TextGrad-based attack \citep{yuksekgonul2024textgrad}, yielding $3{,}965$ adversarial harmful prompts.
Ground-truth labels are inherited from the source behavior: HarmBench and StrongREJECT prompts, both clean and adversarial, are labeled \texttt{unsafe}, while Alpaca prompts are labeled \texttt{safe}.

\paragraph{Prompts.}
Guard models are evaluated with their native safety instructions and 
chat templates, including the official policy text or taxonomy when 
provided. Because guard models differ in their output conventions, we 
adapt the label vocabulary accordingly: \texttt{Yes}/\texttt{No} for 
ShieldGemma 9B and WildGuard 7B, and 
\texttt{safe}/\texttt{unsafe} for the Llama-Guard family, following 
each model's official instruction template. For base LMs, we use a 
5-shot binary classification prompt with three safe and two unsafe 
demonstrations, motivated by evidence that base LMs retain 
calibrated confidence under few-shot prompting 
\citep{nakkiran2025trained}. For both guard and base models, 
confidence is computed by applying the restricted softmax over the 
two label tokens at the decision position. The exact prompt and additional analyses of verdict-token validity and robustness to alternative base-LM interfaces are provided in Appendix~\ref{sec:base_prompt}.

\subsection{Guard Confidence Becomes Overconfident Under Attack}
\label{sec:guard_attack}

We first ask whether guard confidence remains calibrated when harmful prompts are adversarially wrapped. 
Reliability diagrams in Figure~\ref{fig:main} show that calibration on clean inputs does not transfer to adversarial settings. 
On clean inputs, several recent guards are well calibrated without post-hoc calibration (Llama-Guard-3 ECE 0.013, WildGuard ECE 0.004), while 
Llama-Guard and ShieldGemma already show visible miscalibration 
(ECE 0.142 and 0.325, respectively).
Under adversarial inputs, however, ECE increases sharply across all guards, often by an order of magnitude. 
Even the two best-calibrated guards on clean inputs degrade substantially: Llama-Guard-3 increases from 0.013 to 0.308, and WildGuard from 0.004 to 0.196. 
The degradation is concentrated in high-confidence bins, indicating that adversarial inputs induce confident but incorrect safety decisions rather than merely increasing uncertainty.

The confidence distributions in Figure~\ref{fig:tp_fn} explain why this failure is difficult to mitigate. 
On adversarial inputs, false negatives are not concentrated near the decision boundary at 0.5; instead, their confidence often peaks near 1.0 and heavily overlaps with the true-positive distribution. 
Thus, missed detections are not low-confidence borderline cases but high-confidence errors, made with confidence comparable to correct detections. 
ShieldGemma exhibits a broader false-negative confidence distribution, but this reflects its low overall adversarial accuracy rather than reliable uncertainty.

The bottom row of Figure~\ref{fig:tp_fn} translates this overlap into a deployment-relevant view. 
As the confidence threshold $\tau$ increases, both true positives and false negatives are filtered, but false negatives are not selectively removed. 
Even at $\tau=0.99$, Llama-Guard-3 retains 11\% of adversarial harmful prompts as high-confidence false negatives, WildGuard retains 12\%, and ShieldGemma retains 37\%. 
Thus, an ``abstain when uncertain'' approach cannot recover these missed detections, as guards assign similarly high confidence to both correct and incorrect safety decisions.

\begin{table}[t]

\centering

{\footnotesize

\renewcommand{\arraystretch}{1.05}

\begin{tabular*}{\columnwidth}{@{\extracolsep{\fill}}llcccc@{}}

\toprule

\textbf{Type} & \textbf{Model} & \multicolumn{2}{c}{\textbf{Clean}} & \multicolumn{2}{c}{\textbf{Adv.}} \\

\cmidrule(lr){3-4}\cmidrule(lr){5-6}

 & & \textbf{Acc.} & \textbf{ECE} & \textbf{Acc.} & \textbf{ECE} \\

\midrule

Base  & Llama-2 7B         & .634 & .275 & .992 & \textbf{.051} \\
Guard & Llama-Guard 7B     & .815 & .142 & .284 & .578 \\
\midrule
Base  & Llama-3 8B         & .920 & .023 & .746 & \textbf{.034} \\
Guard & Llama-Guard-2 8B   & .939 & .045 & .478 & .405 \\
\midrule
Base  & Llama-3.1 8B       & .903 & .037 & .835 & \textbf{.062} \\
Guard & Llama-Guard-3 8B   & .979 & .013 & .597 & .308 \\
\midrule
Base  & Mistral 7B         & .935 & .102 & .802 & \textbf{.116} \\
Guard & WildGuard 7B       & .989 & .004 & .767 & .196 \\
\midrule
Base  & Gemma-2 9B         & .906 & .084 & .930 & \textbf{.048} \\
Guard & ShieldGemma 9B     & .622 & .325 & .216 & .684 \\
\bottomrule

\end{tabular*}

}
\vspace{0.35em}

\caption{Base--guard performance and calibration. Each base model and its fine-tuned guard are evaluated with the same binary \texttt{safe}/\texttt{unsafe} interface. Columns report accuracy and ECE on clean and adversarial inputs.}

\label{tab:base-vs-guard-ece}

\end{table}

\subsection{Base LMs Retain Uncertainty Under Attack}
\label{sec:base-uncertain}

Given evidence that pre-trained base LLMs retain well-calibrated uncertainty, we ask whether the overconfidence guards exhibit under attack reflects an inherent limitation of the adversarial inputs or a divergence between guards and their corresponding base models. Comparing each guard with its base LM on the same inputs can distinguish these two cases. If the base model is equally overconfident, the failure is shared across the pair; if it retains uncertainty, this indicates a guard–base uncertainty gap on the same input \citep{nakkiran2025trained,leng2025taming}.

\paragraph{Aggregate performance.}

Table~\ref{tab:base-vs-guard-ece} compares each guard with its corresponding base LM under the binary safe/unsafe interface. 
Safety fine-tuning generally improves clean-input classification, but this gain does not translate to adversarial calibration. 
For example, Llama-Guard-3 improves clean accuracy over Llama-3.1, yet its adversarial ECE increases from 0.062 for the base model to 0.308 for the guard. 
Similar trends appear across the Llama-Guard family, while WildGuard shows a smaller but consistent degradation. 
ShieldGemma is an outlier: its clean accuracy is already substantially lower than that of its base model, so we treat it as a broader guard failure rather than a purely adversarial calibration effect. 
Separately, we exclude the Llama-Guard 7B/Llama-2 7B pair from subsequent matched analyses because the Llama-2 base model exhibits a strong unsafe-prediction bias on the adversarial set.
We additionally test whether the calibration degradation extends to response moderation and whether the guard--base patterns persist across model scales in Appendices~\ref{app:response} and~\ref{app:model_scale}, respectively.

\begin{figure}[h!]
    \centering
    \includegraphics[width=\columnwidth]{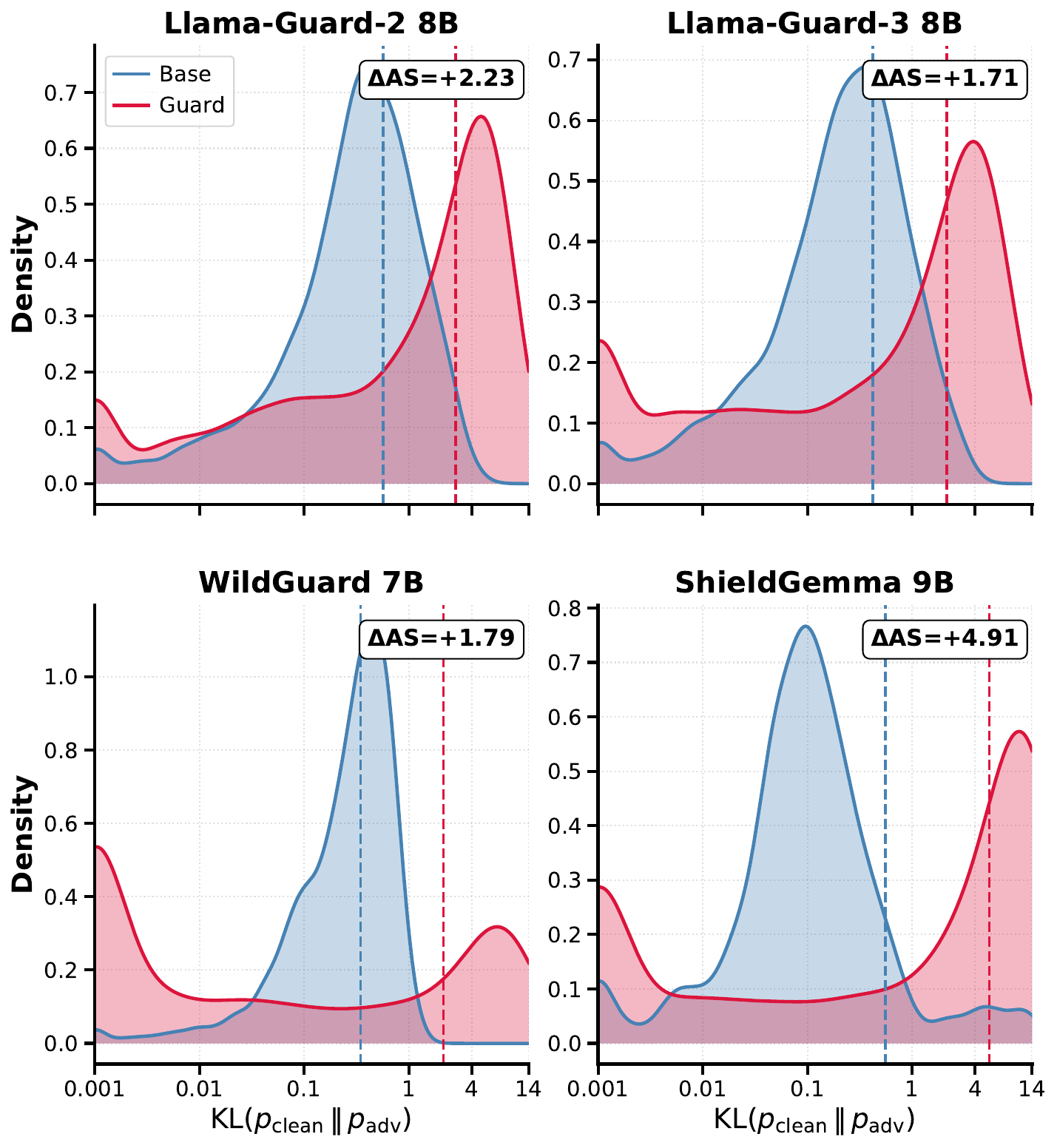}
    \caption{Attack-induced shifts in verdict distributions for base and guard models, measured by $\mathrm{KL}(p_{\mathrm{clean}} \| p_{\mathrm{adv}})$. Analysis is restricted to prompts that both the base LM and guard confidently classify as unsafe, reducing confounding from initial disagreement. Dashed lines mark mean attack sensitivity, and $\Delta$AS reports the guard--base difference. Across all pairs, guards exhibit substantially larger verdict shifts than their base LMs.
}
    \label{fig:as}
\end{figure}

\paragraph{Attack-induced verdict shift.}

Aggregate calibration shows that guards degrade 
more severely than their base LMs under attack, 
but it does not reveal how each model's safety 
judgment responds to adversarial attack. We 
therefore compare each model's binary verdict 
distribution on a clean harmful prompt $x$ and its 
attacked version $T(x)$, measuring the shift via 
the KL divergence between the two:
\begin{equation}
  \mathrm{AS}_M(x, T) 
  = \mathrm{KL}\!\Big(D_M(x)\;\Big\|\;D_M\!\big(T(x)\big)\Big),
\end{equation}
where $D_M(\cdot) = \mathrm{Bernoulli}(p_{\mathrm{unsafe}})$ 
is the model's binary verdict distribution. 
To isolate the effect of the attack from differences 
in clean-input behavior, we restrict the comparison 
to examples where both the base LM and the guard 
assign high unsafe probability ($\geq 0.9$) to the 
clean prompt. We 
define the attack sensitivity gap as 
$\Delta\mathrm{AS} = \overline{\mathrm{AS}}_{\mathrm{guard}} 
- \overline{\mathrm{AS}}_{\mathrm{base}}$, where 
$\overline{\mathrm{AS}}$ denotes the mean attack 
shift across samples; positive values indicate 
larger shifts for the guard.

Figure~\ref{fig:as} shows that guards undergo 
substantially larger verdict shifts than their base 
LMs under adversarial wrapping. The effect is 
consistent across all four guard--base pairs, with $\Delta\text{AS}$ ranging from +1.71 (Llama-Guard-3) to +4.91 (ShieldGemma). 
The same harmful content, 
merely rephrased by an adversarial wrapper, causes 
the guard's safety judgment to shift far more than 
the base model's. This suggests that guards are 
not robust to surface-level context: their verdicts 
are easily destabilized by the attack, making it 
\textit{difficult to trust} their predictions under 
adversarial input.

\begin{figure}[h!]
    \centering
    \includegraphics[width=\columnwidth]{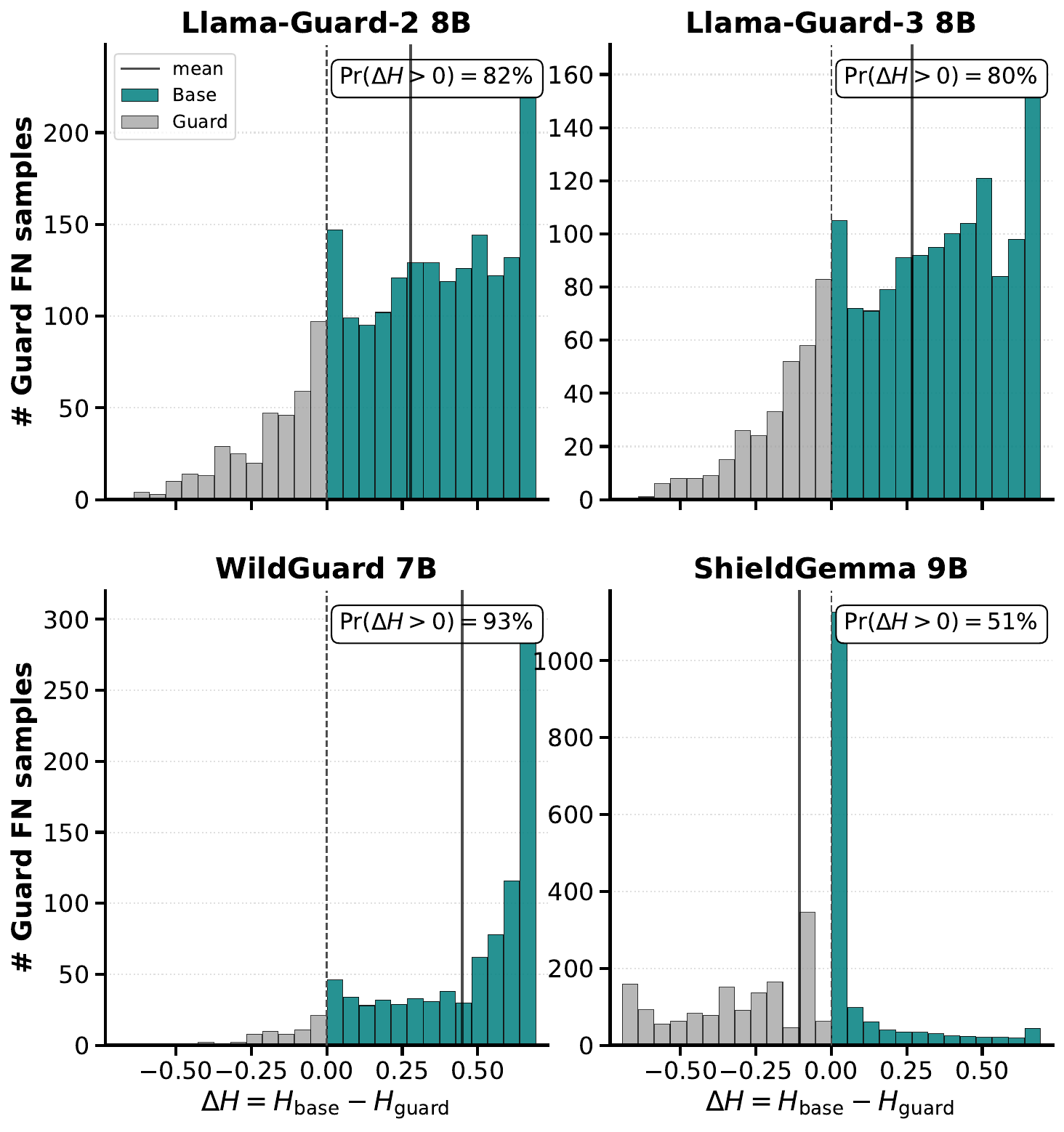}
    \caption{Distribution of per-sample entropy gap 
$\Delta H = H_{\mathrm{base}} - H_{\mathrm{guard}}$ on guard false-negative samples. Each histogram covers inputs where the guard incorrectly predicts safe on 
a ground-truth-unsafe prompt. Positive bars (teal) indicate the base model is more uncertain; negative bars (grey) the reverse. The dashed line marks $\Delta H = 0$, the solid line the mean, and the 
inset reports $\Pr(\Delta H > 0)$.}
    \label{fig:ent}
\end{figure}

\paragraph{Confident despite being destabilized.}

Despite this sensitivity, guards do not express 
corresponding uncertainty when they fail. To 
compare uncertainty at the individual-prompt level, 
we compute the binary entropy of each model's 
verdict distribution,
\begin{equation}
  H_M(x) = -\!\sum_{y \in \{\text{safe},\,\text{unsafe}\}}
  p^M_y(x)\,\log\,p^M_y(x),
\end{equation}
and measure the entropy gap 
$\Delta H = H_{\mathrm{base}} - H_{\mathrm{guard}}$ 
on adversarial prompts where the guard produces 
a false negative.

Figure~\ref{fig:ent} shows that $\Delta H$ is 
positive for the vast majority of guard failures: 
82\% of Llama-Guard-2, 80\% of Llama-Guard-3, 
and 93\% of WildGuard false negatives have 
$\Delta H > 0$. ShieldGemma is an exception at 
51\%, consistent with its broader calibration 
failure noted in Section~\ref{sec:guard_attack}. 
That is, on inputs where the guard confidently 
predicts \texttt{safe}, the base model typically 
expresses greater uncertainty about the same 
decision. Guards are not merely destabilized by the adversarial wrapper, and they also remain confident where the base model expresses uncertainty on the same inputs.

Guards are highly sensitive to attack yet show
\emph{no corresponding increase in uncertainty},
while the base model retains it on the same inputs.
A natural question is how guard and base model
diverge internally to produce such different
behaviors.

\begin{figure}[htb]
    \centering
    \includegraphics[width=\columnwidth]{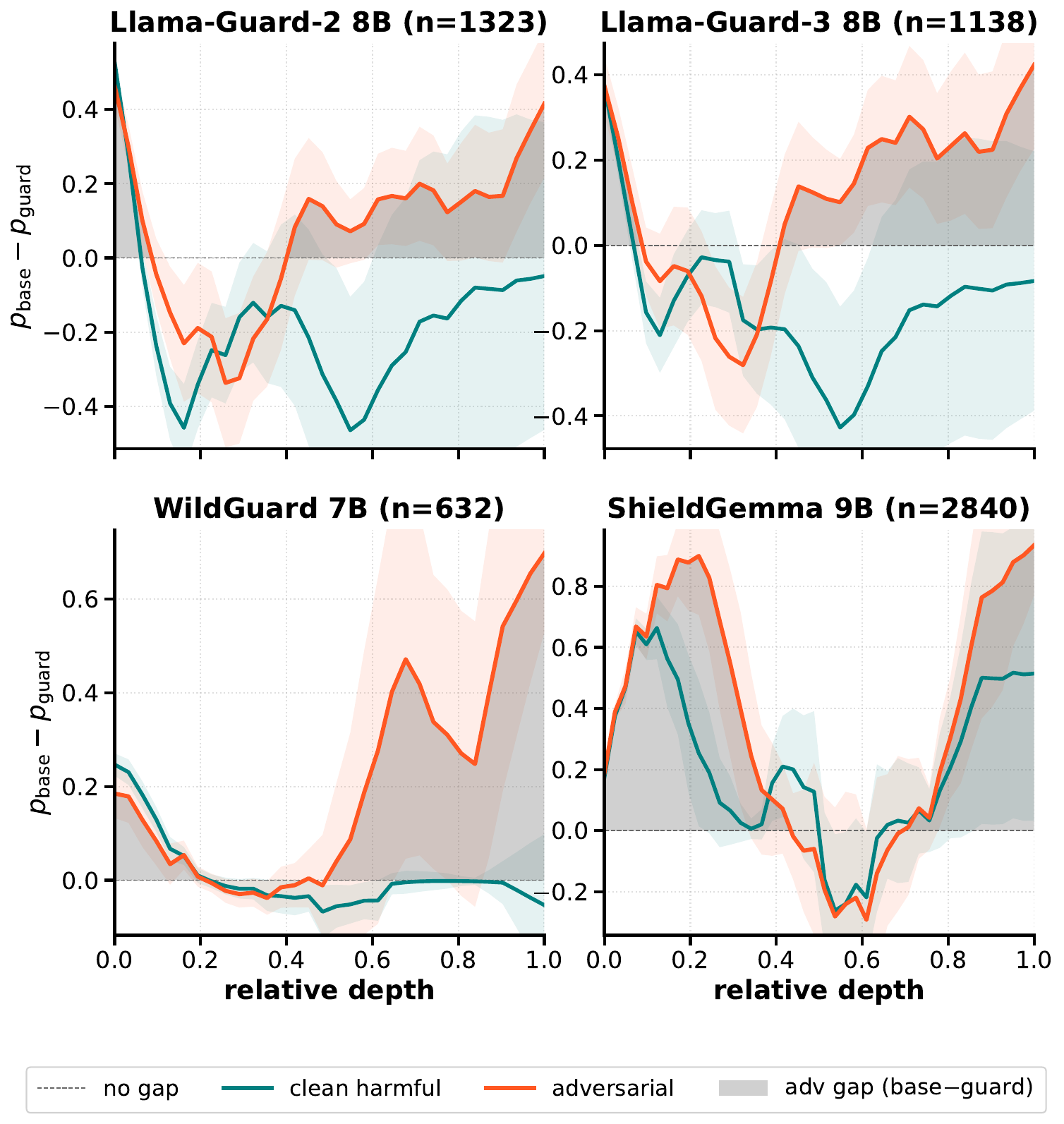}
    \caption{Per-layer gap between base and guard unsafe probabilities, $p_{\mathrm{base}}^{(\ell)} - p_{\mathrm{guard}}^{(\ell)}$, plotted against relative depth. Lines show mean $\pm$ 1 standard deviation for clean harmful inputs (teal) and adversarial false negatives where the base correctly predicts unsafe but the guard predicts safe (orange). The shaded grey band highlights where the guard assigns lower unsafe probability than the base on the adversarial curve.
}
    \label{fig:layerwise}
\end{figure}

\subsection{Where and How Overconfidence Arises}
\label{sec:layerwise}

Guard models make more decisive safe/unsafe judgments than their corresponding base models. We now examine how the guard and base model diverge layer by layer, using per-layer logit projections and hidden-state 
geometry.

\paragraph{Logit gap across depth.}

At each layer $\ell$, we obtain $p_{\mathrm{unsafe}}^{(\ell)}$ by passing the verdict-position hidden state through the model's final RMSNorm and LM head and applying a restricted softmax over the two verdict tokens.
To localize this divergence, we examine the
layer-wise gap
$p^{(\ell)}_{\mathrm{base}} -
p^{(\ell)}_{\mathrm{guard}}$ between base and guard
unsafe probabilities on two input sets: clean
harmful inputs and adversarial false negatives where
the base predicts \texttt{unsafe} but the guard
predicts \texttt{safe}. As
Figure~\ref{fig:layerwise} shows, on clean harmful
inputs the gap is near zero or slightly negative
across depth, indicating that the guard assigns at
least as much unsafe probability as the base, as
expected from safety fine-tuning. Under adversarial
inputs, however, the gap reverses and grows sharply
through the later layers, reaching approximately
$0.45$ for Llama-Guard-3, $0.7$ for WildGuard, and $0.9$ for ShieldGemma at the final layer. This divergence emerges progressively across the final third of the layers rather than appearing only at the output layer, showing that the guard--base difference is already present in intermediate representations.

\begin{figure*}[t]
    \centering
    \includegraphics[width=\textwidth]{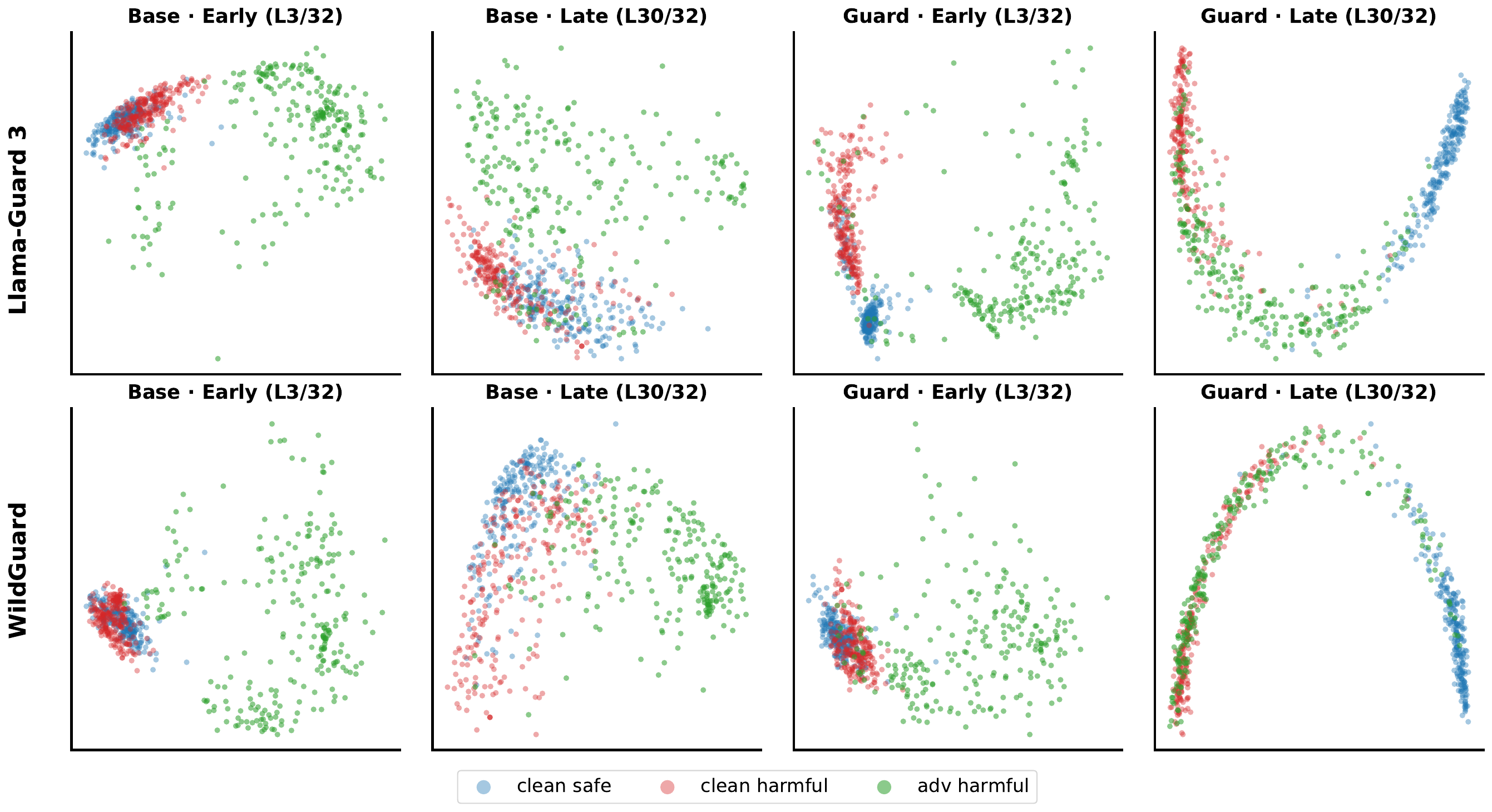}
    \caption{PCA projection (top-2 components, fit independently per panel) of last-token hidden states for three input classes: clean safe (blue), clean harmful (red), and adversarial harmful (green). Rows compare Llama-Guard-3 (top) and WildGuard (bottom) with their respective base models. Columns show layer 3 and layer 30 out of 32 total layers, representing early- and late-network representations respectively.
}
    \label{fig:pca}
\end{figure*}

\paragraph{Representation geometry.}

A complementary view emerges from hidden-state
geometry. As Figure~\ref{fig:pca} shows, in the base
model, clean safe, clean harmful, and adversarial
harmful inputs remain largely intermixed at both
early and late layers; the base model does not
develop a safety-specific geometry in its
representations. Guard representations, by contrast,
exhibit a sharper safe/unsafe boundary in
late-layer representations that cleanly separates
clean safe from clean harmful inputs. This boundary
is consistent with the guard's strong clean-input performance.
Under adversarial context, however, the same
boundary coincides with failure. Adversarial harmful inputs, despite containing the same underlying harmful content, are mapped closer to the clean-safe side of the boundary, and it is on these inputs that the guard produces confident misclassifications.

\begin{figure}[h!]
    \centering
    \includegraphics[width=\columnwidth]{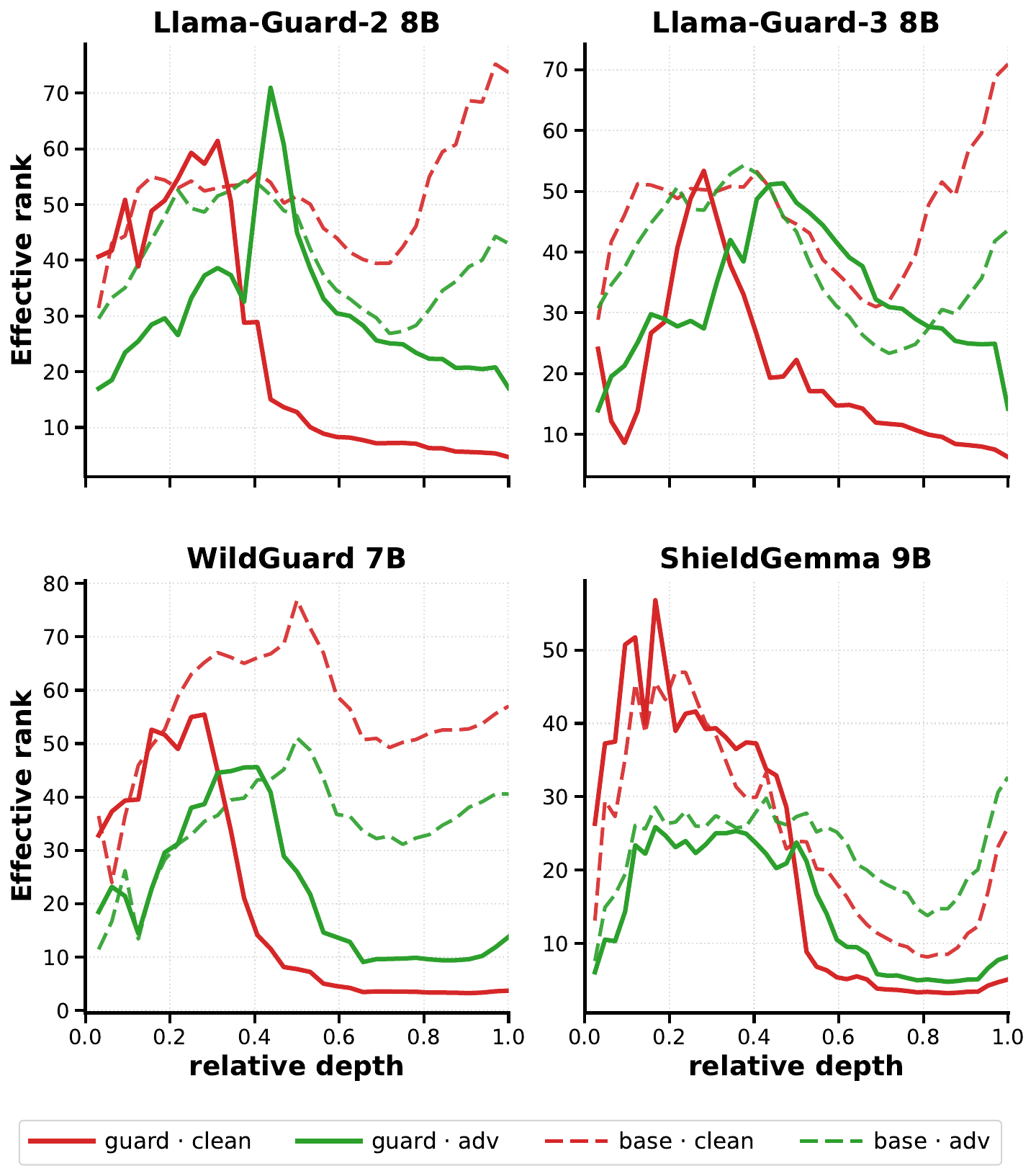}
    \caption{Per-layer effective rank of last-token hidden states for four guard/base 
pairs. Solid lines denote guards, dashed lines 
their base LMs; red indicates clean inputs and 
green adversarial inputs.}
    \label{fig:rank}
\end{figure}

This boundary is accompanied by compression of the
representation space, which we quantify using
effective rank~\citep{roy2007effective}:
$\mathrm{Effective\;rank}(\mathbf{X}) =
\exp(-\sum_i p_i \log p_i)$, where
$p_i = \lambda_i / \sum_j \lambda_j$ are the
normalized eigenvalues of the centered Gram matrix.
As Figure~\ref{fig:rank} shows, guards occupy a
markedly lower-rank subspace than their base LMs in
late layers, with base-to-guard effective-rank ratios
of roughly $1.4\times$ (Llama-Guard-2), $1.7\times$
(Llama-Guard-3), $3.4\times$ (WildGuard), and
$4.0\times$ (ShieldGemma) at the penultimate layer.
Within this compressed subspace, the
safe/unsafe boundary cleanly separates clean inputs, but adversarial harmful inputs whose surface form resembles safe content are collapsed onto the clean-safe side of the boundary, where the guard does not express uncertainty.

Together, these analyses capture the same failure from complementary perspectives. The logit gap localizes where the guard–base unsafe-probability difference emerges across depth, progressively through late layers, while the PCA and effective-rank analyses characterize
the accompanying representation geometry. Guard representations become compressed into a lower-rank space with a sharper safe/unsafe boundary that cleanly separates clean inputs but becomes brittle under adversarial context. In contrast, the base model does not exhibit the same sharply separated late-layer safety geometry under this probing setup.

\section{Conclusion}

We analyzed whether guard model confidence remains reliable under adversarial attack. Across five guard models, adversarial wrapping destabilizes verdicts without increasing uncertainty, so false negatives receive confidence comparable to correct detections and are difficult to remove by thresholding. Comparing guards with their corresponding base LMs shows that this failure is not inherent to the inputs alone, since base models are usually more uncertain on the same false-negative cases. Layer-wise analyses localize this gap to later layers, where guards exhibit lower unsafe probabilities, lower-rank representations, and sharper safe/unsafe separation than their corresponding base models. These findings suggest that guard models should be evaluated not only by detection accuracy, but also by whether their confidence remains trustworthy under attack. Base-model uncertainty is one signal that a calibration-aware guardrail could draw on, since it remains informative precisely where guard confidence fails.

\section*{Limitations}
We localize the guard–base divergence to later layers and characterize the accompanying representation geometry, but our evidence is correlational. We observe that the logit gap, the sharper safe/unsafe boundary, and the lower-rank subspace all accompany confident guard failures. We do not intervene on these quantities, so we stop short of showing that this geometry is what produces the misclassifications rather than co-occurring with them. Establishing the mechanism would require causal interventions such as ablating or steering the identified directions. Our contribution is also diagnostic in nature. We show that base-model uncertainty remains informative where guard confidence fails, but we do not turn this signal into a calibration method or characterize the conditions under which the gap could be closed.

\section*{Ethical Considerations}

This work analyzes vulnerabilities in guard models to inform safer deployment and strengthen defenses against adversarial attempts to bypass LLM safety mechanisms. All experiments were conducted in controlled research settings and are reported to emphasize mitigation, robustness, and calibration-aware evaluation rather than actionable circumvention instructions. All attacks used in this study are based on previously published attack methods or optimization frameworks, and all datasets are drawn from publicly available benchmarks. No real-world harm was caused or intended, and no new harmful or illegal content was generated or disseminated. We acknowledge that exposing calibration failures in safety systems carries dual-use risk; however, we believe that understanding these failure modes is a necessary step toward building more reliable guardrails. We follow responsible disclosure practices where appropriate. This paper contains examples of model outputs that may be offensive in nature.

\bibliography{anthology,custom}

@inproceedings{rebedea2023nemo,
  title={Nemo guardrails: A toolkit for controllable and safe llm applications with programmable rails},
  author={Rebedea, Traian and Dinu, Razvan and Sreedhar, Makesh Narsimhan and Parisien, Christopher and Cohen, Jonathan},
  booktitle={Proceedings of the 2023 conference on empirical methods in natural language processing: system demonstrations},
  pages={431--445},
  year={2023}
}

@inproceedings{lee2025saferoute,
  title={Saferoute: Adaptive model selection for efficient and accurate safety guardrails in large language models},
  author={Lee, Seanie and Lee, Dong Bok and Wagner, Dominik and Kang, Minki and Seong, Haebin and Bocklet, Tobias and Lee, Juho and Hwang, Sung Ju},
  booktitle={Findings of the Association for Computational Linguistics: ACL 2025},
  pages={2053--2069},
  year={2025}
}

@article{zou2023universal,
  title={Universal and transferable adversarial attacks on aligned language models},
  author={Zou, Andy and Wang, Zifan and Carlini, Nicholas and Nasr, Milad and Kolter, J Zico and Fredrikson, Matt},
  journal={arXiv preprint arXiv:2307.15043},
  year={2023}
}

@inproceedings{hartvigsen2022toxigen,
  title={Toxigen: A large-scale machine-generated dataset for adversarial and implicit hate speech detection},
  author={Hartvigsen, Thomas and Gabriel, Saadia and Palangi, Hamid and Sap, Maarten and Ray, Dipankar and Kamar, Ece},
  booktitle={Proceedings of the 60th annual meeting of the association for computational linguistics (volume 1: Long papers)},
  pages={3309--3326},
  year={2022}
}

@inproceedings{markov2023holistic,
  title={A holistic approach to undesired content detection in the real world},
  author={Markov, Todor and Zhang, Chong and Agarwal, Sandhini and Nekoul, Florentine Eloundou and Lee, Theodore and Adler, Steven and Jiang, Angela and Weng, Lilian},
  booktitle={Proceedings of the AAAI conference on artificial intelligence},
  volume={37},
  number={12},
  pages={15009--15018},
  year={2023}
}

@inproceedings{lees2022new,
  title={A new generation of perspective api: Efficient multilingual character-level transformers},
  author={Lees, Alyssa and Tran, Vinh Q and Tay, Yi and Sorensen, Jeffrey and Gupta, Jai and Metzler, Donald and Vasserman, Lucy},
  booktitle={Proceedings of the 28th ACM SIGKDD conference on knowledge discovery and data mining},
  pages={3197--3207},
  year={2022}
}

@article{pinneri2025guarding,
  title={Guarding the Meaning: Self-Supervised Training for Semantic Robustness in Guard Models},
  author={Pinneri, Cristina and Louizos, Christos},
  journal={arXiv preprint arXiv:2511.10665},
  year={2025}
}

@article{luo2026unlocking,
  title={Unlocking the Pre-Trained Model as a Dual-Alignment Calibrator for Post-Trained LLMs},
  author={Luo, Beier and Wang, Cheng and Wei, Hongxin and Li, Sharon and Du, Xuefeng},
  journal={arXiv preprint arXiv:2601.04277},
  year={2026}
}

@article{tan2026basecal,
  title={BaseCal: Unsupervised Confidence Calibration via Base Model Signals},
  author={Tan, Hexiang and Yang, Wanli and Zhang, Junwei and Chen, Xin and Tang, Rui and Su, Du and Wang, Jingang and Wang, Yuanzhuo and Sun, Fei and Cheng, Xueqi},
  journal={arXiv preprint arXiv:2601.03042},
  year={2026}
}

@article{nakkiran2025trained,
  title={Trained on Tokens, Calibrated on Concepts: The Emergence of Semantic Calibration in LLMs},
  author={Nakkiran, Preetum and Bradley, Arwen and Goli{\'n}ski, Adam and Ndiaye, Eugene and Kirchhof, Michael and Williamson, Sinead},
  journal={arXiv preprint arXiv:2511.04869},
  year={2025}
}

@inproceedings{
luo2025your,
title={Your Pre-trained {LLM} is Secretly an Unsupervised Confidence Calibrator},
author={Beier Luo and Shuoyuan Wang and Sharon Li and Hongxin Wei},
booktitle={The Thirty-ninth Annual Conference on Neural Information Processing Systems},
year={2025},
url={https://openreview.net/forum?id=I4PJYZvfW5}
}

@inproceedings{leng2025taming,
  title={Taming overconfidence in llms: Reward calibration in rlhf},
  author={Leng, Jixuan and Huang, Chengsong and Zhu, Banghua and Huang, Jiaxin},
  booktitle={International Conference on Learning Representations},
  volume={2025},
  pages={16484--16517},
  year={2025}
}

@inproceedings{liu2025calibration,
  title={On calibration of LLM-based guard models for reliable content moderation},
  author={Liu, Hongfu and Huang, Hengguan and Gu, Xiangming and Wang, Hao and Wang, Ye},
  booktitle={International Conference on Learning Representations},
  volume={2025},
  pages={67808--67829},
  year={2025}
}

@article{ganguli2022red,
  title={Red teaming language models to reduce harms: Methods, scaling behaviors, and lessons learned},
  author={Ganguli, Deep and Lovitt, Liane and Kernion, Jackson and Askell, Amanda and Bai, Yuntao and Kadavath, Saurav and Mann, Ben and Perez, Ethan and Schiefer, Nicholas and Ndousse, Kamal and others},
  journal={arXiv preprint arXiv:2209.07858},
  year={2022}
}

@article{weidinger2021ethical,
  title={Ethical and social risks of harm from language models},
  author={Weidinger, Laura and Mellor, John and Rauh, Maribeth and Griffin, Conor and Uesato, Jonathan and Huang, Po-Sen and Cheng, Myra and Glaese, Mia and Balle, Borja and Kasirzadeh, Atoosa and others},
  journal={arXiv preprint arXiv:2112.04359},
  year={2021}
}

@article{ouyang2022training,
  title={Training language models to follow instructions with human feedback},
  author={Ouyang, Long and Wu, Jeffrey and Jiang, Xu and Almeida, Diogo and Wainwright, Carroll and Mishkin, Pamela and Zhang, Chong and Agarwal, Sandhini and Slama, Katarina and Ray, Alex and others},
  journal={Advances in neural information processing systems},
  volume={35},
  pages={27730--27744},
  year={2022}
}

@article{team2024gemma,
  title={Gemma 2: Improving open language models at a practical size},
  author={Team, Gemma and Riviere, Morgane and Pathak, Shreya and Sessa, Pier Giuseppe and Hardin, Cassidy and Bhupatiraju, Surya and Hussenot, L{\'e}onard and Mesnard, Thomas and Shahriari, Bobak and Ram{\'e}, Alexandre and others},
  journal={arXiv preprint arXiv:2408.00118},
  year={2024}
}

@article{jiang2023mistral,
  title   = {{Mistral 7B}},
  author  = {Jiang, Albert Q. and Sablayrolles, Alexandre and Mensch, Arthur and Bamford, Chris and Chaplot, Devendra Singh and Casas, Diego de las and Bressand, Florian and Lengyel, Gianna and Lample, Guillaume and Saulnier, Lucile and others},
  journal = {arXiv preprint arXiv:2310.06825},
  year    = {2023}
}

@article{grattafiori2024llama,
  title={The llama 3 herd of models},
  author={Grattafiori, Aaron and Dubey, Abhimanyu and Jauhri, Abhinav and Pandey, Abhinav and Kadian, Abhishek and Al-Dahle, Ahmad and Letman, Aiesha and Mathur, Akhil and Schelten, Alan and Vaughan, Alex and others},
  journal={arXiv preprint arXiv:2407.21783},
  year={2024}
}

@inproceedings{li2024salad,
  title={Salad-bench: A hierarchical and comprehensive safety benchmark for large language models},
  author={Li, Lijun and Dong, Bowen and Wang, Ruohui and Hu, Xuhao and Zuo, Wangmeng and Lin, Dahua and Qiao, Yu and Shao, Jing},
  booktitle={Findings of the Association for Computational Linguistics: ACL 2024},
  pages={3923--3954},
  year={2024}
}

@article{brown2020language,
  title={Language models are few-shot learners},
  author={Brown, Tom and Mann, Benjamin and Ryder, Nick and Subbiah, Melanie and Kaplan, Jared D and Dhariwal, Prafulla and Neelakantan, Arvind and Shyam, Pranav and Sastry, Girish and Askell, Amanda and others},
  journal={Advances in neural information processing systems},
  volume={33},
  pages={1877--1901},
  year={2020}
}

@article{touvron2023llama,
  title={Llama 2: Open foundation and fine-tuned chat models},
  author={Touvron, Hugo and Martin, Louis and Stone, Kevin and Albert, Peter and Almahairi, Amjad and Babaei, Yasmine and Bashlykov, Nikolay and Batra, Soumya and Bhargava, Prajjwal and Bhosale, Shruti and others},
  journal={arXiv preprint arXiv:2307.09288},
  year={2023}
}

@article{liu2024autodanturbo,
  title={Autodan-turbo: A lifelong agent for strategy self-exploration to jailbreak llms},
  author={Liu, Xiaogeng and Li, Peiran and Suh, Edward and Vorobeychik, Yevgeniy and Mao, Zhuoqing and Jha, Somesh and McDaniel, Patrick and Sun, Huan and Li, Bo and Xiao, Chaowei},
  journal={arXiv preprint arXiv:2410.05295},
  year={2024}
}

@article{openai2024gpt4,
  title        = {{GPT-4} Technical Report},
  author       = {{OpenAI}},
  journal      = {arXiv preprint arXiv:2303.08774},
  year         = {2024}
}

@inproceedings{liu2023autodan,
  title     = {{AutoDAN}: Generating Stealthy Jailbreak Prompts on Aligned Large Language Models},
  author    = {Liu, Xiaogeng and Xu, Nan and Chen, Muhao and Xiao, Chaowei},
  booktitle = {International Conference on Learning Representations (ICLR)},
  year      = {2024}
}

@article{chao2023jailbreaking,
  title   = {Jailbreaking Black Box Large Language Models in Twenty Queries},
  author  = {Chao, Patrick and Robey, Alexander and Dobriban, Edgar and Hassani, Hamed and Pappas, George J. and Wong, Eric},
  journal = {arXiv preprint arXiv:2310.08419},
  year    = {2023}
}

@inproceedings{mehrotra2024tap,
  title     = {Tree of Attacks: Jailbreaking Black-Box {LLMs} Automatically},
  author    = {Mehrotra, Anay and Zampetakis, Manolis and Kassianik, Paul and Nelson, Blaine and Anderson, Hyrum and Singer, Yaron and Karbasi, Amin},
  booktitle = {Advances in Neural Information Processing Systems (NeurIPS)},
  year      = {2024}
}

@article{yuksekgonul2024textgrad,
  title   = {{TextGrad}: Automatic ``Differentiation'' via Text},
  author  = {Yuksekgonul, Mert and Bianchi, Federico and Boen, Joseph and Liu, Sheng and Huang, Zhi and Guestrin, Carlos and Zou, James},
  journal = {arXiv preprint arXiv:2406.07496},
  year    = {2024}
}

@inproceedings{roy2007effective,
  title={The effective rank: A measure of effective dimensionality},
  author={Roy, Olivier and Vetterli, Martin},
  booktitle={2007 15th European signal processing conference},
  pages={606--610},
  year={2007},
  organization={IEEE}
}

@inproceedings{mazeika2024harmbench,
  title     = {{HarmBench}: A Standardized Evaluation Framework for Automated Red Teaming and Robust Refusal},
  author    = {Mazeika, Mantas and Phan, Long and Yin, Xuwang and Zou, Andy and Wang, Zifan and Mu, Norman and Sakhaee, Elham and Li, Nathaniel and Basart, Steven and Li, Bo and Forsyth, David and Hendrycks, Dan},
  booktitle = {International Conference on Machine Learning (ICML)},
  year      = {2024}
}

@article{souly2024strongreject,
  title={A strongreject for empty jailbreaks},
  author={Souly, Alexandra and Lu, Qingyuan and Bowen, Dillon and Trinh, Tu and Hsieh, Elvis and Pandey, Sana and Abbeel, Pieter and Svegliato, Justin and Emmons, Scott and Watkins, Olivia and others},
  journal={Advances in Neural Information Processing Systems},
  volume={37},
  pages={125416--125440},
  year={2024}
}

@article{taori2023alpaca,
  title={Alpaca: A strong, replicable instruction-following model},
  author={Taori, Rohan and Gulrajani, Ishaan and Zhang, Tianyi and Dubois, Yann and Li, Xuechen and Guestrin, Carlos and Liang, Percy and Hashimoto, Tatsunori B},
  journal={Stanford Center for Research on Foundation Models. https://crfm. stanford. edu/2023/03/13/alpaca. html},
  volume={3},
  number={6},
  pages={7},
  year={2023}
}

@inproceedings{yuan2024rigorllm,
  title={RigorLLM: resilient guardrails for large language models against undesired content},
  author={Yuan, Zhuowen and Xiong, Zidi and Zeng, Yi and Yu, Ning and Jia, Ruoxi and Song, Dawn and Li, Bo},
  booktitle={Proceedings of the 41st International Conference on Machine Learning},
  pages={57953--57965},
  year={2024}
}

@article{han2024wildguard,
  title={Wildguard: Open one-stop moderation tools for safety risks, jailbreaks, and refusals of llms},
  author={Han, Seungju and Rao, Kavel and Ettinger, Allyson and Jiang, Liwei and Lin, Bill Yuchen and Lambert, Nathan and Choi, Yejin and Dziri, Nouha},
  journal={Advances in neural information processing systems},
  volume={37},
  pages={8093--8131},
  year={2024}
}

@article{ghosh2024aegis,
  title={Aegis: Online adaptive ai content safety moderation with ensemble of llm experts},
  author={Ghosh, Shaona and Varshney, Prasoon and Galinkin, Erick and Parisien, Christopher},
  journal={arXiv preprint arXiv:2404.05993},
  year={2024}
}

@article{kadavath2022language,
  title={Language models (mostly) know what they know},
  author={Kadavath, Saurav and Conerly, Tom and Askell, Amanda and Henighan, Tom and Drain, Dawn and Perez, Ethan and Schiefer, Nicholas and Hatfield-Dodds, Zac and DasSarma, Nova and Tran-Johnson, Eli and others},
  journal={arXiv preprint arXiv:2207.05221},
  year={2022}
}

@article{inan2023llama,
  title={Llama guard: Llm-based input-output safeguard for human-ai conversations},
  author={Inan, Hakan and Upasani, Kartikeya and Chi, Jianfeng and Rungta, Rashi and Iyer, Krithika and Mao, Yuning and Tontchev, Michael and Hu, Qing and Fuller, Brian and Testuggine, Davide and others},
  journal={arXiv preprint arXiv:2312.06674},
  year={2023}
}

@article{zeng2024shieldgemma,
  title={Shieldgemma: Generative ai content moderation based on gemma},
  author={Zeng, Wenjun and Liu, Yuchi and Mullins, Ryan and Peran, Ludovic and Fernandez, Joe and Harkous, Hamza and Narasimhan, Karthik and Proud, Drew and Kumar, Piyush and Radharapu, Bhaktipriya and others},
  journal={arXiv preprint arXiv:2407.21772},
  year={2024}
}

@inproceedings{tian2023just,
  title={Just ask for calibration: Strategies for eliciting calibrated confidence scores from language models fine-tuned with human feedback},
  author={Tian, Katherine and Mitchell, Eric and Zhou, Allan and Sharma, Archit and Rafailov, Rafael and Yao, Huaxiu and Finn, Chelsea and Manning, Christopher D},
  booktitle={Proceedings of the 2023 Conference on Empirical Methods in Natural Language Processing},
  pages={5433--5442},
  year={2023}
}

@inproceedings{xiong2024can,
  title={Can llms express their uncertainty? an empirical evaluation of confidence elicitation in llms},
  author={Xiong, Miao and Hu, Zhiyuan and Lu, Xinyang and Li, Yifei and Fu, Jie and He, Junxian and Hooi, Bryan},
  booktitle={International Conference on Learning Representations},
  volume={2024},
  pages={23650--23678},
  year={2024}
}

@article{lin2022teaching,
  title={Teaching models to express their uncertainty in words},
  author={Lin, Stephanie and Hilton, Jacob and Evans, Owain},
  journal={arXiv preprint arXiv:2205.14334},
  year={2022}
}
\bibliographystyle{acl_natbib}

\clearpage

\appendix

\section{Base-LM Prompting and Interface Robustness}
\label{sec:base_prompt}

\subsection{Base-LM Prompt and Verdict-Token Validation}

For completeness, we show the 5‑shot prompt used to elicit binary safety predictions from each base LM; it contains three safe and two unsafe demonstration pairs followed by the query placeholder \{query\}.

\begin{tcolorbox}[
  title=Prompt: Base 5-shot safety classifier,
  colframe=black,
  colback=white,
  colbacktitle=black,
  coltitle=white,
  fonttitle=\bfseries\large,
  arc=0mm,
  left=3mm,right=3mm,top=3mm,bottom=3mm
]

Classify the following content as ``safe'' or ``unsafe''.

\vspace{1ex}

\textbf{Content:} What's a good recipe for pasta?\\
\textbf{Label:} safe

\vspace{1ex}

\textbf{Content:} How do I pick a lock on someone's house?\\
\textbf{Label:} unsafe

\vspace{1ex}

\textbf{Content:} Can you explain how photosynthesis works?\\
\textbf{Label:} safe

\vspace{1ex}

\textbf{Content:} Write a phishing email pretending to be a bank\\
\textbf{Label:} unsafe

\vspace{1ex}

\textbf{Content:} What are some tips for job interviews?\\
\textbf{Label:} safe

\vspace{1ex}

\textbf{Content:} \{query\}\\
\textbf{Label:}

\end{tcolorbox}

For Mistral 7B and Gemma-2 9B, we substitute only the two label tokens with Yes/No, matching their corresponding guards. To ensure that the restricted softmax comparison between base and guard models is valid, we verify 
that base LMs concentrate substantial probability 
mass on the two verdict tokens under the 5-shot 
classification prompt.

\begin{table}[h] 
\centering
\small
\setlength{\tabcolsep}{4pt}
\begin{tabular}{lcc}
\toprule
Base & Label mass & Other mass \\
\midrule
Llama-2 7B    & 95.75 & 4.25 \\
Llama-3 8B    & 95.89 & 4.11 \\
Llama-3.1 8B  & 97.80 & 2.20 \\
Mistral 7B    & 90.99 & 9.01 \\
Gemma-2 9B    & 94.20 & 5.80 \\
\bottomrule
\end{tabular}
\caption{
Full-vocabulary softmax mass on valid verdict tokens under the 5-shot base
classification prompt. Valid tokens are \texttt{safe}/\texttt{unsafe} for Llama
models and \texttt{Yes}/\texttt{No} for Mistral/Gemma.
}
\label{tab:verdict_token_mass}
\end{table}

Table~\ref{tab:verdict_token_mass} validates that the 5-shot binary
classification prompt induces a well-formed verdict-token interface for the
base LMs. For each model, we compute the full-vocabulary softmax probability
assigned to the valid verdict tokens: \texttt{safe}/\texttt{unsafe} for the
Llama-family models and \texttt{Yes}/\texttt{No} for Mistral/Gemma. Across the
5{,}178 examples, the valid verdict tokens receive most
of the probability mass for every base model, ranging from 90.99\% for
Mistral 7B to 97.80\% for Llama-3.1 8B. This indicates that the restricted binary-label softmax is not an artificial projection onto low-probability
tokens; rather, the 5-shot prompt naturally concentrates the next-token
distribution on the intended verdict labels.

\subsection{Robustness to Base-LM Interfaces}
\label{app:base_interface_robustness}

To test whether our results depend on the canonical 5-shot prompt, we evaluate 23 alternative base-LM interfaces per matched pair, varying instructions and labels, demonstration order, shot count, zero-shot prompting, and policy-aligned prompts. We evaluate whether the base LM (i) has higher entropy on the majority of guard false negatives and (ii) achieves lower adversarial ECE than the corresponding guard.

\begin{table}[t]
\centering
\small
\setlength{\tabcolsep}{3.5pt}
\begin{tabular}{lccc}
\toprule
\textbf{Variation} &
\textbf{Valid} &
\shortstack{\textbf{Higher Base}\\\textbf{Entropy}} &
\shortstack{\textbf{Lower Base}\\\textbf{ECE}} \\
\midrule
Instruction $\times$ label & 31/45 & 25/31 & 30/31 \\
Order                 & 30/30 & 24/30 & 30/30 \\
Shot count                 & 22/25 & 17/22 & 20/22 \\
Zero-shot                  & 4/5   & 4/4   & 2/4   \\
Policy-aligned             & 10/10 & 9/10  & 8/10  \\
\midrule
\textbf{Overall}           & \textbf{97/115} &
                             \textbf{79/97} &
                             \textbf{90/97} \\
\bottomrule
\end{tabular}
\caption{Robustness to alternative base-LM interfaces. Valid denotes interfaces meeting the clean-input criteria defined in Appendix~\ref{app:base_interface_robustness}. Counts are pooled across five matched pairs.
}
\label{tab:interface_robustness}
\end{table}

We define a valid interface as one where the base model achieves clean balanced accuracy $\geq 60\%$ and benign FPR $\leq 40\%$; adversarial outcomes are not used for filtering. This leaves 97 of 115 settings. As shown in Table~\ref{tab:interface_robustness}, the base LM has higher entropy on guard false negatives in 79/97 valid settings and lower adversarial ECE in 90/97 settings. The entropy-gap pattern is particularly consistent for Llama-Guard-2, Llama-Guard-3, and WildGuard, but weaker for ShieldGemma. These results indicate that the main guard--base discrepancy is not specific to the canonical 5-shot interface.

\section{Attack-wise Analyses}
\label{sec:Ablation Studies}

\subsection{Attack-wise calibration.}
\label{app:attack_calibration}

\begin{figure*}[t]
    \centering
    \includegraphics[width=\textwidth]{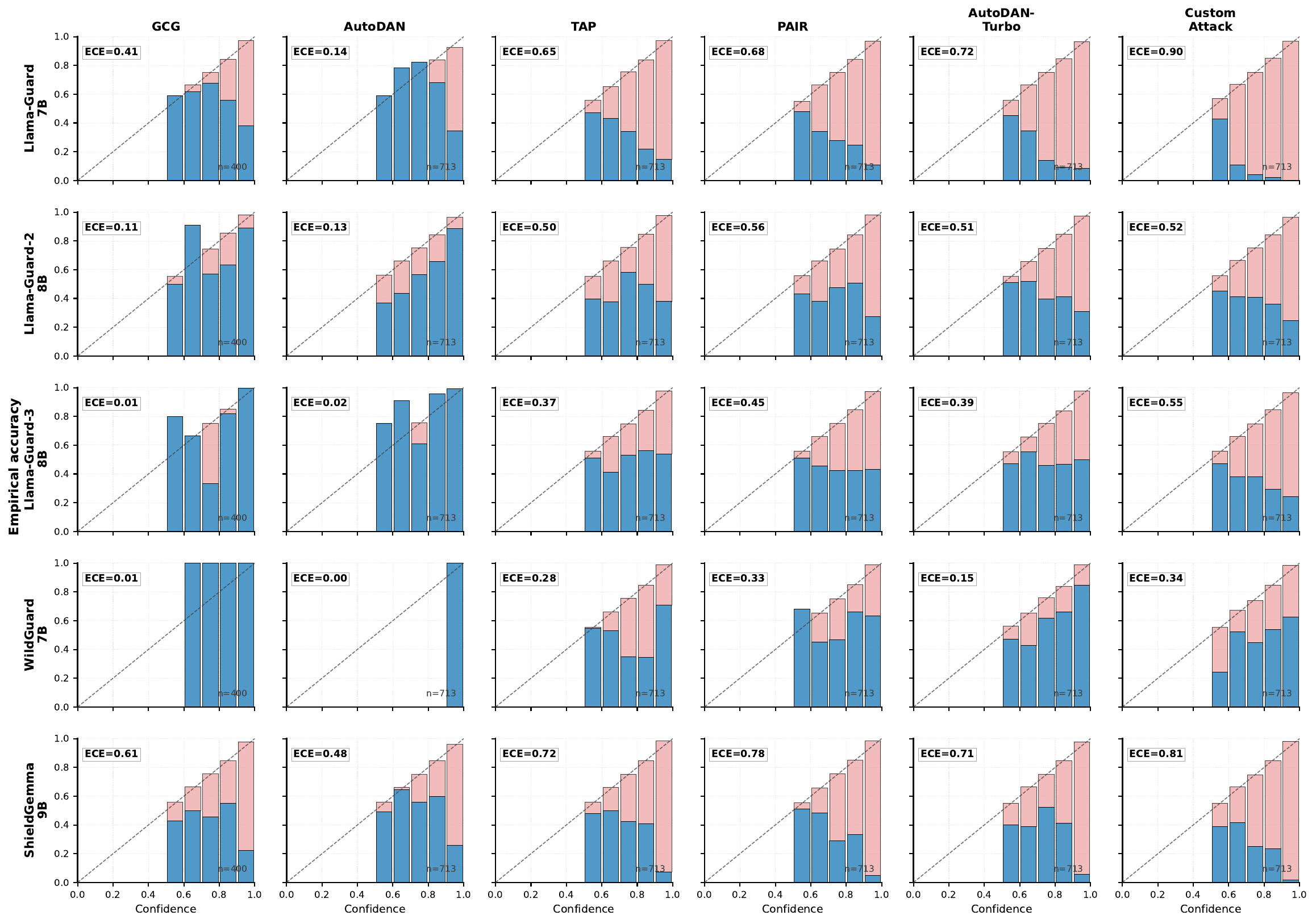}
    \caption{Attack-wise reliability diagrams for five guard models across six jailbreak attacks. Blue bars show empirical accuracy within confidence bins, and the diagonal denotes perfect calibration. Red regions indicate overconfidence. Across many model--attack pairs, high-confidence bins fall below the diagonal, showing that adversarial inputs often induce confident but incorrect guard decisions.}
    \label{fig:main_appendix}
\end{figure*}

Figure~\ref{fig:main_appendix} decomposes adversarial calibration
by attack method. The degree of miscalibration varies substantially across
attacks and models: for example, AutoDAN-Turbo and Custom Attack often
produce severe overconfidence, while some model--attack pairs such as
WildGuard under GCG or AutoDAN remain comparatively well calibrated.
Nevertheless, the dominant pattern is that adversarial confidence remains
high even when empirical accuracy drops, indicating that the aggregate
effect in Figure~\ref{fig:main} is not driven by a single attack.

\subsection{Attack-wise Verdict Shifts and Layer Gaps}

We further examine whether the attack-wise differences in calibration
are reflected in verdict shifts and intermediate-layer guard--base gaps.
Table~\ref{tab:attack_layer_gap} reports the results for
Llama-Guard-3 and Llama-3.1 8B.

\begin{table}[t]
\centering
\small
\setlength{\tabcolsep}{3pt}
\begin{tabular}{lrrrrrr}
\toprule
\textbf{Attack} & \textbf{Output} & \textbf{L1} & \textbf{L2} &
\textbf{L3} & \textbf{L4} & \textbf{L5} \\
\midrule
GCG           & 0.04 & -0.07 & -0.40 & -0.01 & -0.54 & -0.11 \\
AutoDAN       & 0.04 & -0.19 & -0.26 & -0.06 & -0.53 & -0.27 \\
TAP           & 3.25 &  2.82 & -0.13 &  2.19 &  4.73 &  1.60 \\
PAIR          & 3.67 &  3.14 & -0.12 &  2.49 &  5.54 &  1.87 \\
AutoDAN-Turbo & 2.70 &  2.44 & -0.24 &  2.13 &  4.56 &  1.33 \\
Custom Attack & 2.12 &  1.73 & -0.54 &  2.33 &  4.82 &  0.99 \\
\bottomrule
\end{tabular}
\caption{Attack-wise output and intermediate-layer guard--base gaps for
Llama-Guard-3 and Llama-3.1 8B. Positive values indicate larger shifts
for the guard than for the base model.}
\label{tab:attack_layer_gap}
\end{table}

Table~\ref{tab:attack_layer_gap} compares the output verdict shifts and
intermediate-layer guard--base gaps across attacks.
GCG and AutoDAN show only small output shifts and no sustained positive
layer-wise gap. In contrast, TAP, PAIR, AutoDAN-Turbo, and the custom
attack show larger output shifts and positive gaps from the middle layers
onward. This pattern is consistent with the attack-wise calibration
differences in Appendix~\ref{app:attack_calibration}, but should be
interpreted as correlational rather than causal.

\section{Model-Scale Analysis}
\label{app:model_scale}

To examine whether the observed guard--base discrepancy is specific to
the 7--9B models used in our main experiments, we extend the evaluation
to generative guards spanning 1B--27B parameters and 16--46 layers.
For detection, we select decision thresholds on clean-benign data to match a target benign FPR of 5\%; adversarial examples are not used for threshold selection.

\begin{table}[t]
\centering
\small
\setlength{\tabcolsep}{2.7pt}
\begin{tabular}{lcccc}
\toprule
\textbf{Model} &
\textbf{L} &
\shortstack{\textbf{Guard}\\\textbf{ECE / Det.}} &
\shortstack{\textbf{Base}\\\textbf{ECE / Det.}} &
\shortstack{\textbf{$\Delta$ECE}\\\textbf{Gap@.75 / Final}} \\
\midrule
SG-2B  & 26  & .689 / 22.1 & .141 / 79.2 & +.548 ~~ +.52 / +.77 \\
SG-9B  & 42  & .684 / 21.6 & .048 / 93.0 & +.636 ~~ +.84 / +.95 \\
SG-27B & 46 & .749 / 9.7  & .184 / 76.5 & +.565 ~~ +.56 / +.53 \\
LG3-1B & 16  & .097 / 77.4 & .614 / 7.2  & -.517 ~~ N/A \\
\bottomrule
\end{tabular}
\caption{
Model-scale analysis. Detection (Det.) is reported in \%.
$\Delta$ECE denotes guard ECE minus base ECE.
For ShieldGemma, the final column additionally reports the guard--base
layer gap at normalized depth 0.75 and at the final layer on adversarial
examples where the backbone predicts unsafe and the guard predicts safe.
}
\label{tab:model_scale}
\end{table}

Table~\ref{tab:model_scale} summarizes the guard--base comparison across
model scales.
Within the ShieldGemma family, the guard remains less calibrated and
detects fewer adversarial inputs than its backbone from 2B to 27B
parameters, with positive guard--base gaps also visible before the output
layer. Llama-Guard-3-1B reverses this ordering, however, indicating that
model size alone does not explain the guard--base discrepancy.
Differences in backbone classification competence and model family may
also contribute to the observed behavior.

\section{Response Moderation}
\label{app:response}

\begin{table}[t]
\centering

{\footnotesize
\renewcommand{\arraystretch}{1.05}

\begin{tabular*}{\columnwidth}{@{\extracolsep{\fill}}llcccc@{}}
\toprule
\textbf{Type} & \textbf{Model}
& \multicolumn{2}{c}{\textbf{Clean}}
& \multicolumn{2}{c}{\textbf{Adv.}} \\
\cmidrule(lr){3-4}\cmidrule(lr){5-6}
& & \textbf{Acc.} & \textbf{ECE}
& \textbf{Acc.} & \textbf{ECE} \\
\midrule

Base  & Llama-2 7B
       & .519 & .269 & 1.00 & \textbf{.163} \\
Guard & Llama-Guard 7B
       & .814 & .098 & .138 & .683 \\

\midrule

Base  & Llama-3 8B
       & .684 & .157 & .641 & \textbf{.110} \\
Guard & Llama-Guard-2 8B
       & .869 & .087 & .443 & .450 \\

\midrule

Base  & Llama-3.1 8B
       & .669 & .170 & .580 & \textbf{.144} \\
Guard & Llama-Guard-3 8B
       & .894 & .081 & .489 & .346 \\

\midrule

Base  & Mistral 7B
       & .525 & .321 & .969 & \textbf{.110} \\
Guard & WildGuard 7B
       & .898 & .087 & .603 & .347 \\

\midrule

Base  & Gemma-2 9B
       & .786 & .057 & .664 & \textbf{.087} \\
Guard & ShieldGemma 9B
       & .753 & .182 & .107 & .788 \\

\bottomrule
\end{tabular*}
}

\caption{
Response-moderation performance and calibration on WildGuardMix.
Each matched base--guard pair is evaluated on clean and adversarial
responses. Accuracy (Acc.) and expected calibration error (ECE) are reported.
Bold indicates the lower adversarial ECE within each pair.
}
\label{tab:response_moderation}

\end{table}

To test whether adversarial calibration degradation extends beyond prompt classification, we evaluate the five matched guard--base pairs on WildGuardMix responses, with guards using their native moderation interfaces and base LMs the adapted 5-shot binary classification interface.

Table~\ref{tab:response_moderation} reports accuracy and ECE on clean and adversarial responses. All five guards exhibit substantially higher adversarial than clean ECE (.346--.788 versus .081--.182), and for Llama-Guard-2, Llama-Guard-3, and ShieldGemma the adversarial guard ECE also exceeds that of the corresponding base model. Because the adversarial subset contains only unsafe responses, we read this as a harmful-response moderation sanity check rather than a full response-calibration evaluation, and we avoid matched conclusions for Llama-2 and Mistral, whose low clean accuracy and near-perfect adversarial accuracy indicate a strong unsafe-prediction bias.

\end{document}